\documentclass[a4paper,twoside,11pt]{article}

\usepackage[USenglish]{babel} 
\usepackage[T1]{fontenc}

\usepackage{graphicx}
\usepackage[labelsep=endash, figurename=Figure, tablename=Table, textfont=it]{caption}
\usepackage{subcaption}
\usepackage{float}  
\usepackage{hyperref}
\usepackage{multicol}
\usepackage{multirow}
\usepackage{amsmath}
\usepackage{siunitx}

\usepackage{fancyhdr}
\usepackage[top=2.5cm, bottom=2.5cm, left=2.5cm, right=2.5cm]{geometry} 
\usepackage[usenames,dvipsnames]{pstricks}
\usepackage[round,comma,authoryear]{natbib}
\usepackage{titling}
\newif\ifarxiv
\arxivtrue

\definecolor{darkblue}{rgb}{0,0,0.5}
\hypersetup{colorlinks=true,linkcolor=black,citecolor=darkblue,urlcolor=darkblue} 

\makeatletter
\let\ps@plain\ps@fancy
\makeatother

\begin{document}


\title{\bf Systematic Analyses of Reinforcement Learning Controllers\\ in  Signalized Urban Corridors}
\author{Xiaofei Song$^{a, *}$, Kerstin Eder$^{a}$, Jonathan Lawry$^{a}$ and R. Eddie Wilson$^{a}$}
\date{\today}

\pretitle{\centering\Large}
\posttitle{\par\vspace{1ex}}

\preauthor{\centering}
\postauthor{\par\vspace{1ex}
$^{a}$ University of Bristol, Bristol, UK\\
xiaofei.song@bristol.ac.uk, 
Kerstin.Eder@bristol.ac.uk,
J.Lawry@bristol.ac.uk,
RE.Wilson@bristol.ac.uk
\\
$^{*}$ Corresponding author

\vspace{1ex}\it
Extended abstract submitted for presentation at the $6^{th}$ Symposium on Management of Future Motorway and Urban Traffic Systems\\
September 14-15, 2026, Naples, Italy\\
\vspace{1ex}

}

\maketitle
\vspace{-1cm}
\noindent\rule{\textwidth}{0.5pt}\vspace{0cm}

\fancypagestyle{firststyle}{
\lhead[]{}
\rhead[]{}
\lfoot[MFTS26]{MFTS26}
\rfoot[Original abstract submittal]{Original abstract submittal}
\cfoot[]{}
}
\thispagestyle{firststyle}

\pagestyle{fancy}
\fancyhead{}
\fancyfoot{}
\renewcommand{\headrulewidth}{0pt}
\renewcommand{\footrulewidth}{0pt}
\setlength{\headheight}{15pt}
\lhead{Xiaofei Song, Kerstin Eder, Jonathan Lawry and R. Eddie Wilson}
\rhead[\thepage]{\thepage}
\lfoot[MFTS26]{MFTS26}
\rfoot[Original abstract submittal]{Original abstract submittal}
\cfoot[]{}


\section{Introduction}
Reinforcement learning (RL) can directly learn signal timing strategies from traffic state observations, generally outperforming traditional traffic signal control methods in simulation~\citep{wei2019survey}. However, most studies evaluate performance under a single demand setting, without checking how different controllers compare to each other as demand is varied~\citep{zhao2024survey}. In traffic networks, the ability of a signal controller to maintain stable traffic conditions depends on whether the traffic demand can be effectively served by the controller. When demand exceeds the service capacity of the controller, queues grow persistently and the system becomes unstable, leading to spillback and network congestion. In this case, direct comparison of average travel times (ATTs) is not meaningful. Rather, we argue that the performance comparison should only be conducted under traffic demand conditions where all controllers are stable. To address this limitation, we have adopted the concept of the \emph{capacity region} in our previous work~\citep{previouswork}, defined as the set of traffic demand combinations under which a controller maintains stable queue dynamics. 

In this work, we extend our systematic capacity region perspective to multi-junction traffic networks, focussing on the special case of an urban corridor network. In particular, we train and evaluate centralized, fully decentralized, and parameter-sharing decentralized RL controllers, and compare their capacity regions and ATTs together with a classical baseline MaxPressure controller \citep{varaiya2013max}. Further, we show how the parameter-sharing controller may be generalised to be deployed on a larger network than it was originally trained on. In this setting, we show some initial findings that suggest that even though the junctions are not formally coordinated, traffic may self-organise into `green waves'.

\section{Methodology}
\begin{figure}
\includegraphics[width=\linewidth]{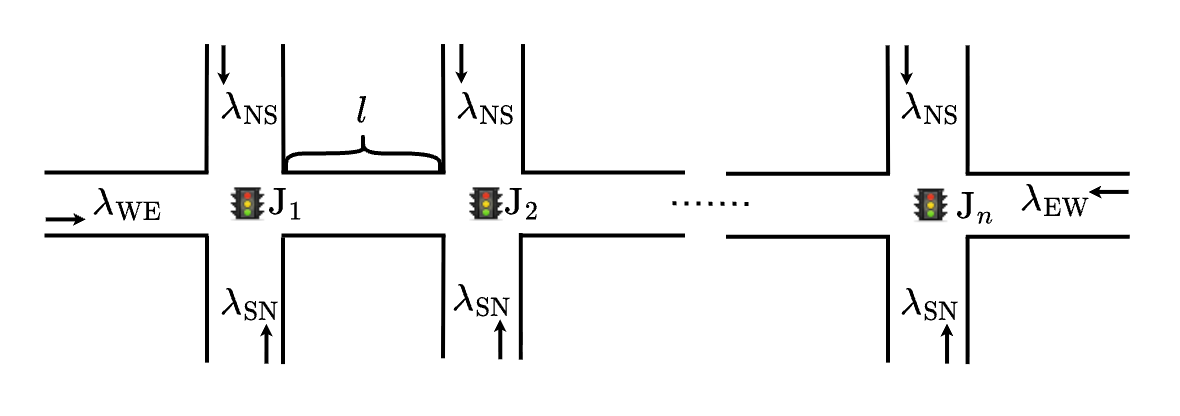}
\vspace{-0.4in}
    \caption{Multi-junction corridor simulation set-up}
    \label{fig:setup}
\end{figure}

We use SUMO~\citep{sumo} to simulate an urban corridor network with $n$ signalized junctions $\mathrm{J}_i$ with $i=1,2,\ldots,n$, equally spaced at inter-junction distance $l$,
see Fig.~\ref{fig:setup}. Vehicles arrive according to independent Poisson processes of differing rates $\lambda$ at the entry links, each of which has a backup queue with unlimited capacity to store unmet traffic demand when the network becomes congested. Statistics of the backup queues are processed to determine whether a given simulation is within capacity or not \citep{previouswork}. All roads have one lane per direction, and all movements at the junctions are straight-on only. The North-South / South-North side-roads share the same traffic demands. Further, we suppose $\lambda_\mathrm{NS}=\lambda_\mathrm{SN}$ throughout, and initially, $\lambda_\mathrm{WE}=\lambda_\mathrm{EW}$, to give a two-dimensional demand space.

Each junction has two phases, alternatively serving the NS/SN and WE/EW movements.
The junctions execute (potentially different) signal control decisions simultaneously every 17\,s, when, either a junction switches phase --- in which case there is a 2\,s amber interval followed by 15\,s of the new phase --- or the current phase is maintained for a further 17\,s, until the next decision. This discrete-time decision structure is adopted as the simplest way to implement RL controllers.
 
We use the MaxPressure controller as a baseline,
which selects signal phases based on queue length imbalances. For RL control, we apply Proximal Policy Optimization (PPO)~\citep{schulman2017proximal}, a widely used policy-gradient method known for its stable training performance.  For PPO, we trained and evaluated three architectures: (i) Centralized PPO: a single controller observes the entire traffic state and outputs signal decisions for all junctions~\citep{liu2025vehicle}.
(ii) Fully decentralized (FD) PPO: Each junction is controlled by an independent controller that observes the traffic state local to its junction and learns its own policy.
(iii) Parameter-sharing (PS) decentralized PPO: a special case of fully decentralized PPO, in which we require that the trained control policy is identical for each junction, enabling decentralized execution with shared learning~\citep{yu2022surprising}.
 
For the RL cases, traffic signal control is formulated as a Markov decision process. The state is defined as the traffic densities on incoming lanes. The action space is binary: the controller either maintains the current phase or switches to the competing phase. The reward is defined as the negative queue length. For decentralized control, each junction receives a local reward equal to the negative sum of queue lengths at that junction. In contrast, the centralized controller receives a global reward defined as the negative total queue length across the network.

The RL controllers are trained under a single demand set-up with $\lambda_{\mathrm{NS}}=\lambda_{\mathrm{WE}}=700$ veh/h, close to the capacity limit of the MaxPressure controller.  The hypothesis is that training close to the capacity limit might produce controllers that can generalize well to different demand set-ups.
 
The RL controllers are trained through 500 simulations of 10,000\,s until convergence. The hightest reward
policy is selected and then evaluated across a set of different demand configurations $(\lambda_{\mathrm{NS}}, \lambda_{\mathrm{WE}})$, to estimate capacity regions and compare ATTs.
 
\section{Results}
\begin{figure}[h]
    \centering
     \begin{minipage}{0.54\textwidth}
         
    \begin{subfigure}[p]{0.49\linewidth}
        \centering
        \includegraphics[width=\linewidth]{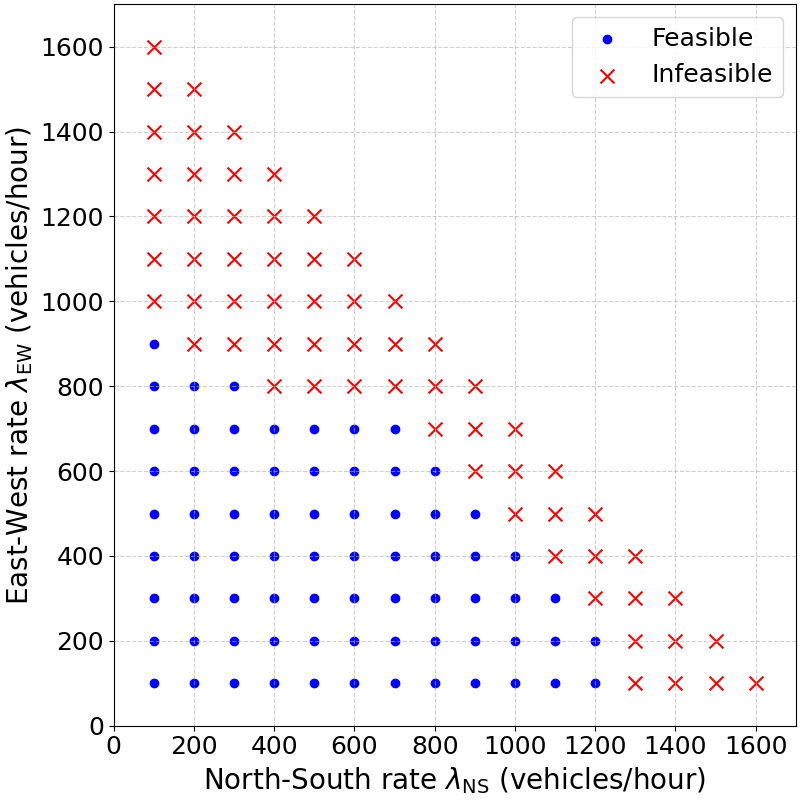}
        \caption{maxPressure}
        \label{fig:cap-mp}
    \end{subfigure}
    \hfill
    \begin{subfigure}[p]{0.49\linewidth}
        \centering  
        \includegraphics[width=\linewidth]{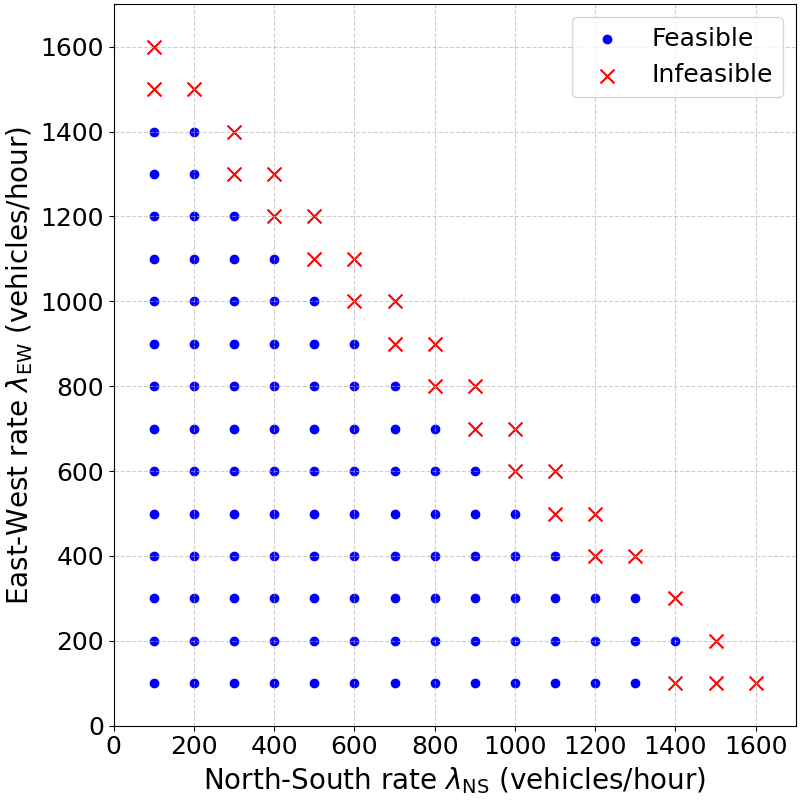}
        \caption{Centralized PPO}
        \label{fig:cap-cen}
    \end{subfigure}
    \\
    \begin{subfigure}[p]{0.49\linewidth}
        \centering
        \includegraphics[width=\linewidth]{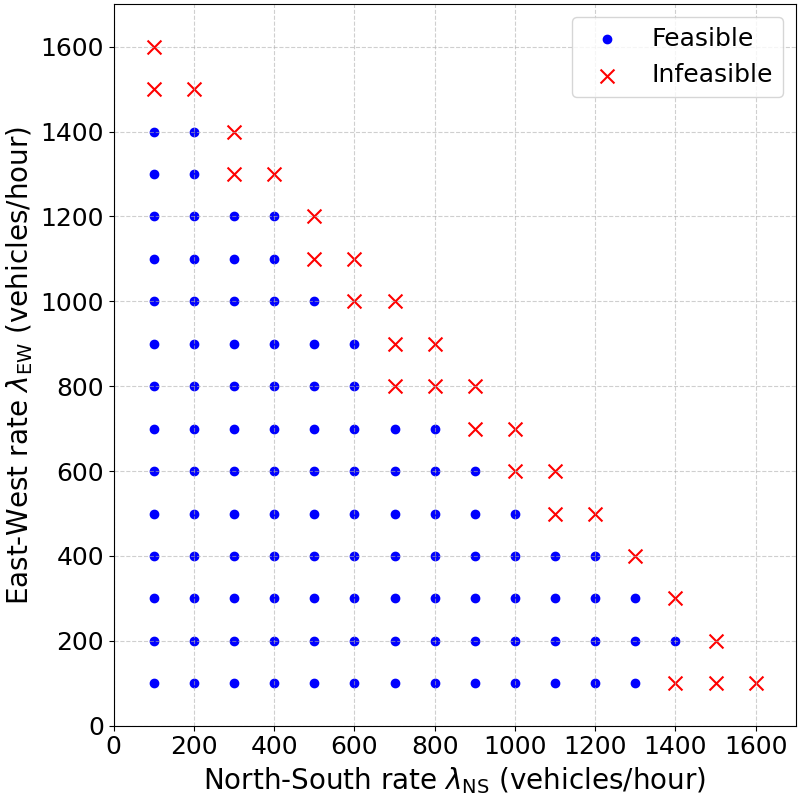}
        \caption{PS decentralized PPO}
        \label{fig:cap-de}
    \end{subfigure}
    \hfill
     \begin{subfigure}[p]{0.49\linewidth}
        \centering
        \includegraphics[width=\linewidth]{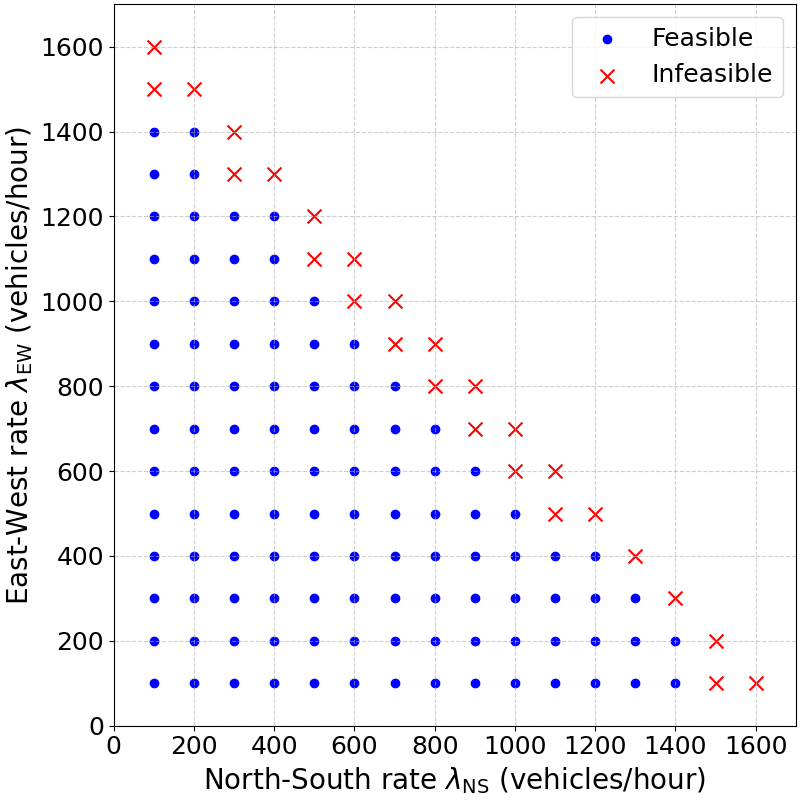}
        \caption{Fully decentralized PPO}
        \label{fig:cap-fullde}
    \end{subfigure} 
     \end{minipage}
    \begin{minipage}{0.45\textwidth}
    \begin{subfigure}[p]{\linewidth}
        \centering
        \small
        \caption{ATT comparison across architectures}
        \label{fig:cap-table}
        \begin{tabular}{S[table-format=4.] S[table-format=4.] S[table-format=4.2] S[table-format=4.2] S[table-format=4.2]}
            \hline
            \multicolumn{1}{c}{\multirow{2}{*}{WE}} &
            \multicolumn{1}{c}{\multirow{2}{*}{NS}} &
            \multicolumn{3}{c}{ATT}  \\ \cline{3-5}
             & &{Centralized} & {PS} & {FD} \\ \hline
100&1300&64.87&65.29&67.39\\
200&1200&68.35&69.26&70.81\\
300&1100&72.99&72.99&74.28\\
400&1000&77.12&77.26&78.51\\
500&900&85.78&84.16&85.07\\
600&800&89.50&90.21&89.99\\
700&700&93.42&92.61&92.43\\
800&600&99.08&98.74&97.57\\
900&500&102.53&107.10&105.59\\
1000&400&105.97&113.44&109.61\\
1100&300&110.80&119.80&115.87\\
1200&200&116.19&126.92&122.10\\
1300&100&123.99&135.24&130.69\\
            \hline
        \end{tabular}
    \end{subfigure}
    \\
    \begin{subfigure}[p]{\linewidth}
        \centering
        \includegraphics[width=\linewidth]{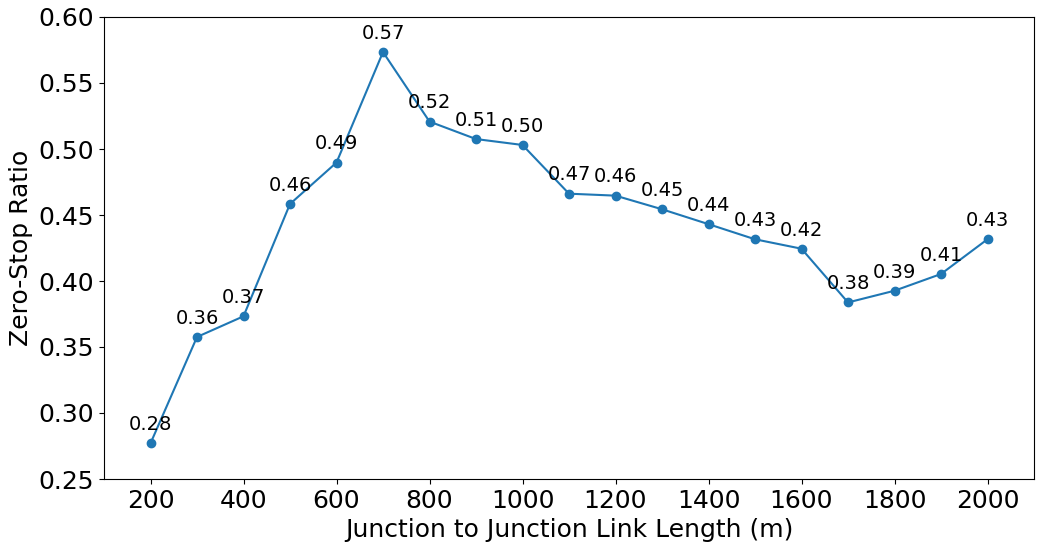}
        \caption{West to East vehicle stop time vs length}
        \label{fig:greenwave}
    \end{subfigure}
   \end{minipage}
    \caption{(a-d) Capacity regions for MaxPressure and RL controllers. (e) Average travel times (ATT). (f) Zero-stop ratios for $n=13$ junctions, suggesting green wave coordination at $l=700\,$m.}
    \label{fig:capacity1}
\end{figure}

Here we focus on results for the case with $n=3$ junctions. Figs.~\ref{fig:capacity1}(a--d) present empirical capacity regions for each of the four signal control strategies. The results show that the three RL controllers operate stably over a substantially larger set of demand configurations than the MaxPressure controller, even though they are trained on a single demand setup.
The capacity region of the MaxPressure controller is notably asymmetric. This can be explained by the corridor structure: vehicles in the WE/EW directions traverse more junctions and are therefore subject to more control decisions, making them more sensitive to congestion and reducing the feasible demand range in that direction --- the RL controllers 
do not suffer from this problem as they are more robust to coordination complexity.
 
Table~\ref{fig:capacity1}(e) reports the ATT for RL controllers for combinations of demand with $\lambda_\mathrm{NS}+\lambda_\mathrm{WE}=1400\,$veh/h --- busy scenarios which are nevertheless within the capacity region of all three controllers.
As expected, the centralized PPO controller consistently achieves the lowest travel times, because it has access to the global traffic state and can also coordinate the junctions directly. However, despite the lack of global information and coordination, the fully decentralized PPO controller performs only slightly worse. A trade-off is involved in the performance of the parameter-sharing decentralized controller. On the one hand, parameter sharing improves the training efficiency, but on the other, the use of a single shared policy is sub-optimal, because the junctions play different roles in the network: $\mathrm{J}_{1,3}$ are at the boundary of the network, whereas $\mathrm{J}_2$ only receives EW/WE flow from the other two junctions. Thus there is decline in performance as the EW/WE flows increase.

However, the benefit of the parameter-sharing decentralized controller is that it can be deployed to the individual junctions in a different network from that which it was trained on. To investigate this, we deployed the controller to each junction on a 13$\times$1 network. We set $\lambda_\mathrm{NS}=\lambda_\mathrm{SN}=100\,$veh/h, $\lambda_\mathrm{WE}=800\,$veh/h, and breaking symmetry, in order to see if a `green-wave' phenomenon  might emerge, we set $\lambda_\mathrm{EW}=0\,$veh/h. Further, we vary the inter-junction distance $l$ from 200\,m to 2000\,m. 
Instead of requiring an entire platoon to traverse all intersections without stopping, we consider that a vehicle experiences a green wave~\citep{greenwave} if it passes
through the corridor without any stopping. Accordingly, we use the \emph{zero-stop ratio}
defined to be the proportion of WE traffic that traverses the entire corridor without a single stop. 
 
As the link length increases, the zero-stop ratio initially rises, reaching a peak value of approximately $0.57$ at $l=700\,$m, which takes approximately 50\,s to drive at the free-flow traffic speed of 50\,km/h. Due to various stochastic effects in the simulation, there is no clean relationship with the signal period. However, the same set-up with $n=1$ junctions yields a zero-stop ratio $\simeq0.93$,
and curiously $0.93^{13}\simeq 0.39, \simeq 0.43$ (the $l\rightarrow\infty$ ratio). Although the controllers are independent, they are coupled via the platoons that pass through them in sequence, and this result suggests that they can coordinate to achieve a theoretical near-optimum performance.

\section{Discussion and Outlook}
Key challenges remain for the practical deployment of RL controllers on real-world traffic networks. Firstly, controllers must perform well when exposed to unexpected demand conditions that are outside their training set --- and we believe our application of capacity regions is the first systematic attempt to understand this kind of robustness. Secondly, controllers should generalize to networks outside their training set, or at least, should not require ground-up training for each new network deployment. The parameter-sharing decentralized controller appears to be a natural solution component for this problem, and our early experiments with it suggest it can facilitate green-wave dynamics, even though there is no direct coordination between junctions. In ongoing work, we are studying the generalisation to $n\times n$ grid-network set-ups.


\begin{small}
\begin{sloppypar} 
\bibliographystyle{authordate1}

\setlength{\bibsep}{0pt}

\bibliography{References}

\end{sloppypar}
\end{small}


\end{document}